# Accuracy analysis of Educational Data Mining using Feature Selection Algorithm

Ali Jaber Almalki

Department of Computer Science, University of Central Florida, Orlando, Florida, United States

***Abstract*** - *Gathering relevant information to predict student academic progress is a tedious task. Due to the large amount of irrelevant data present in databases which provides inaccurate results. Currently, it is not possible to accurately measure and analyze student data because there are too many irrelevant attributes and features in the data. With the help of Educational Data Mining (EDM), the quality of information can be improved. This research demonstrates how EDM helps to measure the accuracy of data using relevant attributes and machine learning algorithms performed. With EDM, irrelevant features are removed without changing the original data. The data set used in this study was taken from Kaggle.com. The results compared on the basis of recall, precision and f-measure to check the accuracy of the student data. The importance of this research is to help improve the quality of educational research by providing more accurate results for researchers.*

## 1 Introduction

Educational Data Mining (EDM) is the growing discipline for exploring unique data and providing the means to gather relevant information from large datasets. EDM provides an efficient approach for exploring educational data and providing accurate information that is specified using different algorithms and classifiers. EDM is used to "mine" the relevant data from large databases therefore providing a higher quality of information to be analyzed. In the educational sector in particular, it is exceedingly difficult to gather information effectively due to the high amount of irrelevant data that is also available in the databases. The tedious work of separating out the irrelevant data from large databases and identifying only the relevant data is not time efficient.

EDM offers a more efficient and time-effective means for collecting target data. EDM helps to represent different computations of data in identical form from different perspectives. In this research evaluation process, EDM shows how accurate data is gathered based only relevant attributes.

EDM removes redundant information and provides precise information resulting in more effective analysis. The main goal of this research is to provide an evaluation process to gather relevant information from student data. Different algorithms and classifiers were used to mine the data. The accuracy results checked on the basis correctly identified instances.

Existing algorithms are not efficient enough to measure the accuracy of the data collected, and due to the lack of accuracy, prediction analysis cannot clearly specify the results of student performance. The process of extracting relevant information does not produce valid results because, during extraction, results are replicated and merged with the results of irrelevant information [6]. Applying the existing algorithms to gather the relevant information only changes the original format of the data and the hidden factor of the educational data can't be effectively explored. EDM aims to provide a means for extracting the hidden factor without changing the original format. EDM provides researchers with the opportunity to explore a large amount of educational data and relevant information can gather to check the academic performance of the student. Extraction of the hidden data helps to provide the necessary information to identify the relevant features of the data. Feature selection algorithm helps to gather the relevant attributes of data and check the accuracy of the data [7]. It explores the hidden aspects of the data and provide identified results to check accuracy of large educational data.

## 2 Literature Review

EDM is the emerging field that investigates different areas of data and explores the hidden information within the data without changing the original information, and then represents the hidden data in an effective manner. Different algorithms have been proposed with different perspectives to improve the accuracy of EDM. For the study of educational data, computational approaches are used to clarify it by dividing it into different clusters and groups.

Different researchers have implemented their efforts to bring out the best possible results from large educational databases. Vasile (2007) suggested the Sequential Pattern algorithm to be used for EDM. This algorithm provides real-time feedback to conclude the results so that further analysis can be made [13]. The Sequential Pattern algorithm uses different mining techniques including clustering, which is classifying the data into diverse groups. The Sequential Pattern algorithm is not able to work on large datasets however, and therefore, generated results are not able to accurately evaluate student performance.

In EDM, different classifiers have been used for the mining of data aimed at specific results. Shafiq *et al.,* (2014) noted the Clustering algorithm to improve EDM [11]. The Clustering algorithm makes groups of classifications so that the evaluation process can go further than with other algorithms and this, in turn, aims to help improve the education system as the results are meaningful. The Correlation-based Feature Subset Selection is a method in which two measures of correlation are used and subsets are further classified. The subsets are evaluated, and classification is made accordingly [4].

Antonio *et al.,* (2004) proposed an algorithm that includes the Consistency-based Feature Subset Selection. Antonio *et al.* made an evaluation matrix on the consistency; the significance is based on the Filter-based method. The results determined that it is difficult to evaluate the data in the specific datasets, and it is more complex due to irrelevant data [2]. Anupama (2011) suggested another EDM algorithm, Gain Ratio Feature Selection algorithm. The Gain Ratio Feature Selection algorithm provides 19 classifiers used to perform a simulation on the data set. The provided classifiers are then used to make the evaluation matrix [1].

Varun (2011) suggested the Naïve Based algorithm. The Naïve Based algorithm is comprised of 11-13 variables. This algorithm helps to build a model in which student performance evaluation can be navigated effectively. The Naïve Based algorithm provides complete results in which the risks, failures, and success of students can all be evaluated [3]. The effectiveness of this information is to help provide the necessary guidelines to improve student productivity and additional counseling programs that should be considered to improve the efficiency of the students [12].

Grafsgaard (2014) developed an algorithm, Information Gain (IG), that identifies the facial expressions and behavior of students. Frustration of students is detected through this algorithm [5]. The IG algorithm generates the Prediction Model which provides insight into student performance, so that, based on the results, educators can work to improve the education system. However, student performance was affected by continuously changing mood swings, making it difficult to obtain meaningful results [8]. Though IG has been used in EDM, the precision value to measure the performance of the student cannot be measured accurately and researchers are therefore unable to evaluate the results effectively. Katrina *et al.* (2015) proposed an algorithm that includes the Correlation Feature Selection. This algorithm includes the Filter Feature algorithm for mining the data to get the specified results [9]. However, by using the results of the classification, the original data is altered.

## 3 Methodology

For the exploration of the educational data, the data collected from the Kaggle.com. The data consists of 480 instances. It is an educational data that contains the information on student performance and marks according to the different study level. The data are taken from the student database and using the WEKA, the performance analysis made to analyze the evaluation process of the student's performance. The goal of this study research is to measure student performance accuracy in terms of correctly classified as per the algorithm performance.

### 3.1 Educational Data Mining Process

Educational Data Mining (EDM) is the process through which techniques and algorithms are applied to evaluate the data on student performance. The results measure the accuracy of the data and correctly identify the instances. For this process, the following algorithms were applied: FS algorithm Cfs Subset Eval, Filtered Attribute Eval, Gain Ratio Attribute Eval, Principal Components, and Relief Attribute Eval. The following classification algorithms were applied: Bayes Net (BN), Jrip, Naïve Bayes (NB), Naïve Bayes Updateable (NBU), MLP, Decision Stump (DS), Simple Logistic (SL), SMO, Decision Table (DT), OneR, REP tree (RepT).

### 3.2 Experimental Setup and Dataset

The dataset was taken from Kaggle.com and WEKA was used to perform the evaluation of the algorithm. The primary aim of study was to identify the best combinations of FS algorithms and classifiers to identify the key student performance factors [10]. For the experimental setup, WEKA was used, and wrapper and filter methods were used for data mining.

# 4 Results and Discussions

The investigation and concluding results in this research focus on the performance of the different classifiers of the machine learning algorithms with the related classifiers. The performance effectiveness of the algorithms was based on precision, recall, and F-measure. Correctly classified algorithms were taken to evaluate the best evaluation process of the algorithm with the calculated mean. The results are shown with the nine algorithms and in nine separate tables, each showing the measure of precision, recall, and F-measure.

## 4.1 CFS Subset Evaluation

CFS Subset evaluation helps to measure the features of the data by considering a separate attribute of the dataset. Predictive capabilities were identified using this algorithm and homogenous features were used in the CFS Subset Evaluation for the selection process.

*Table 1 Evaluation Performance of CFS Subset Evaluation*

| Classification Algorithm | Precision | Recall | F- Measure |
|---|---|---|---|
| NB | 0.892 | 0.892 | 0.893 |
| BN | 0.887 | 0.883 | 0.885 |
| SL | 0.887 | 0.883 | 0.885 |
| MLP | 0.918 | 0.919 | 0.918 |
| NBU | 0.895 | 0.896 | 0.896 |
| SMO | 0.889 | 0.885 | 0.887 |
| Jrip | 0.892 | 0.892 | 0.892 |
| DT | 0.921 | 0.921 | 0.921 |
| DS | 0.868 | 0.821 | 0.830 |
| PART | 0.933 | 0.933 | 0.933 |
| OneR | 0.868 | 0.821 | 0.830 |
| J48 | 0.921 | 0.922 | 0.923 |
| RepT | 0.912 | 0.913 | 0.912 |

Table 1 represents the evaluation process of the CFS Subset Evaluation using precision, recall, and F-measure. In all the classifiers, the CFS algorithm was used as the base algorithm with the different classifiers. However, Decision Stump and One R show slightly lower performance as compared to the other classifiers.

## 4.2 Filtered Attribute Evaluation

Filtered Attribute Evaluation algorithm is accessible in the WEKA plate form and does the evaluation process by selecting features of the data sets. By filtering the attributes of the training data, it provides evaluation results.

*Table 2 Evaluation Performance of Filtered Attribute Evaluation*

| Algorithm Classification | Precision | Recall | F-Measure |
|---|---|---|---|
| NB | 0.899 | 0.896 | 0.897 |
| BN | 0.901 | 0.888 | 0.891 |
| SL | 0.901 | 0.888 | 0.891 |
| MLP | 0.994 | 0.994 | 0.994 |
| NBU | 0.918 | 0.919 | 0.918 |
| SMO | 0.923 | 0.923 | 0.923 |
| Jrip | 0.899 | 0.900 | 0.899 |
| DT | 0.923 | 0.923 | 0.923 |
| DS | 0.868 | 0.821 | 0.830 |
| PART | 0.953 | 0.952 | 0.951 |
| OneR | 0.868 | 0.821 | 0.830 |
| J48 | 0.901 | 0.902 | 0.901 |
| RepT | 0.914 | 0.915 | 0.914 |

Table 2 shows the evaluation performance of the Filtered Attribute Evaluation. The concluding results demonstrate that One R and DS show comparatively low results based on precision, recall, and F-measure, whereas, MLP shows higher results as compared to the other classifiers.

## 4.3 Gain Ratio Attribute

Gain Ratio Attribute algorithm is the non-symmetrical measure that gains information based on attribute selection and evaluates the results accordingly. It takes the common feature of the dataset and relates it with all the other features of the class.

*Table 3 Evaluation Performance of Gain Ratio Attribute*

| Classification Algorithm | Precision | Recall | F- Measure |
|---|---|---|---|
| BN | 0.899 | 0.896 | 0.897 |
| NB | 0.901 | 0.888 | 0.891 |
| NBU | 0.901 | 0.888 | 0.891 |
| MLP | 0.994 | 0.994 | 0.994 |
| SL | 0.918 | 0.919 | 0.918 |
| SMO | 0.923 | 0.923 | 0.923 |
| DT | 0.899 | 0.900 | 0.899 |
| Jrip | 0.923 | 0.923 | 0.923 |
| OneR | 0.868 | 0.821 | 0.830 |
| PART | 0.953 | 0.952 | 0.951 |
| DS | 0.868 | 0.821 | 0.830 |
| J48 | 0.901 | 0.902 | 0.901 |
| RepT | 0.914 | 0.915 | 0.914 |

Table 4 shows the evaluation of the data set using the Gain Ratio Attribute and provides all the results-based classification algorithm. DT, One R and DS show low results and MLP, PART shows high results based on recall, precision, and F-measure.

## 4.4 Principal Components

The Principal Components algorithm helps to decrease the space dimensionality without the decrement of original features of the dataset.

*Table 4 Evaluation Performance of Principal Components*

| Algorithm Classification | Precision | Recall | F-Measure |
|---|---|---|---|
| NB | 0.864 | 0.860 | 0.862 |
| BN | 0.864 | 0.848 | 0.852 |
| SL | 0.864 | 0.848 | 0.852 |
| MLP | 0.988 | 0.988 | 0.988 |
| NBU | 0.905 | 0.906 | 0.905 |
| SMO | 0.905 | 0.906 | 0.906 |
| Jrip | 0.874 | 0.877 | 0.875 |
| DT | 0.913 | 0.915 | 0.914 |
| DS | 0.813 | 0.819 | 0.800 |
| PART | 0.973 | 0.971 | 0.971 |
| OneR | 0.735 | 0.735 | 0.735 |
| J48 | 0.941 | 0.942 | 0.941 |
| RepT | 0.885 | 0.888 | 0.885 |

Using the open data mining tool WEKA, the performance component shown in Table 4 demonstrates the results with 13 classifiers. One R shows low results based on recall, precision, and F-measure, whereas results produced by the MLP show better performance.

## 4.5 Relief Attribute Evaluation

Measuring the significance of the Relief Attribute Evaluation algorithm helps to measure the data statistically. Statistically related attributes are measured using this algorithm. Significance is measured by repeated sampling.

Table 5 shows the evaluation results and provides detailed results of all the classification algorithms by following the base algorithm. DS shows lows results based on precision, recall, and F-measure, whereas MLP shows the best results.

*Table 5 Evaluation Performance of Relief Attribute Evaluation*

| Algorithm Classification | Precision | Recall | F-Measure |
|---|---|---|---|
| NB | 0.899 | 0.896 | 0.897 |
| BN | 0.901 | 0.888 | 0.891 |
| SL | 0.901 | 0.888 | 0.891 |
| MLP | 0.994 | 0.994 | 0.994 |
| NBU | 0.918 | 0.919 | 0.918 |
| SMO | 0.923 | 0.923 | 0.923 |
| Jrip | 0.899 | 0.900 | 0.899 |
| DT | 0.923 | 0.923 | 0.923 |
| DS | 0.868 | 0.821 | 0.830 |
| PART | 0.953 | 0.952 | 0.951 |
| OneR | 0.868 | 0.821 | 0.830 |
| J48 | 0.901 | 0.902 | 0.901 |
| RepT | 0.914 | 0.915 | 0.914 |

## 4.6 Correlation Based Attribute Evaluation

Correlation-based Attribute Evaluation helps to measure the subset features based on hypothesis. This evaluation process helps to correlate and un-correlate with the classifications.

*Table 6 Correlation Based Attribute Evaluation*

| Algorithm Classification | Precision | Recall | F-Measure |
|---|---|---|---|
| NB | 0.899 | 0.896 | 0.897 |
| BN | 0.901 | 0.888 | 0.891 |
| SL | 0.901 | 0.888 | 0.891 |
| MLP | 0.994 | 0.994 | 0.994 |
| NBU | 0.918 | 0.919 | 0.918 |
| SMO | 0.923 | 0.923 | 0.923 |
| Jrip | 0.900 | 0.899 | 0.899 |
| DT | 0.923 | 0.923 | 0.923 |
| DS | 0.868 | 0.821 | 0.830 |
| PART | 0.953 | 0.952 | 0.951 |
| OneR | 0.868 | 0.821 | 0.830 |
| J48 | 0.901 | 0.902 | 0.901 |
| RepT | 0.914 | 0.915 | 0.914 |

Table 6 shows that MLP provides a good correlation with the classification, whereas OneR shows a low correlation process based on recall, precession, and F-measure.

## 4.7 Information Gain Attribute Evaluation

This algorithm helps to measure the information based on the selected class. Attribute worth can be measured effectively using the Information Gain Attribute that depends on the class.

*Table 7 Evaluation Performance of Information Gain Attribute*

| Algorithm Classification | Precision | Recall | F-Measure |
|---|---|---|---|
| NB | 0.899 | 0.896 | 0.897 |
| BN | 0.901 | 0.888 | 0.891 |
| SL | 0.901 | 0.888 | 0.891 |
| MLP | 0.994 | 0.994 | 0.994 |
| NBU | 0.918 | 0.919 | 0.918 |
| SMO | 0.923 | 0.923 | 0.923 |
| Jrip | 0.899 | 0.900 | 0.899 |
| DT | 0.923 | 0.923 | 0.923 |
| DS | 0.868 | 0.821 | 0.830 |
| PART | 0.953 | 0.952 | 0.951 |
| OneR | 0.868 | 0.821 | 0.830 |
| J48 | 0.901 | 0.902 | 0.901 |
| RepT | 0.914 | 0.915 | 0.914 |

Table 7 shows the results of the Information Gian Attribute Evaluation based on precision, recall, and F-measure. MLP classifier performs the best evaluation as compared to the other classifiers.

## 4.8 One R Attribute Evaluation

This helps to measure the worth of the classifier depending on the training dataset. It individually measures the worth of the attribute and provides accurate results.

*Table 8 Evaluation Performance of One R Attribute Evaluation*

| Algorithm Classification | Precision | Recall | F-Measure |
|---|---|---|---|
| NB | 0.899 | 0.896 | 0.897 |
| BN | 0.901 | 0.888 | 0.891 |
| SL | 0.901 | 0.888 | 0.891 |
| MLP | 0.994 | 0.994 | 0.994 |
| NBU | 0.918 | 0.919 | 0.918 |
| SMO | 0.923 | 0.923 | 0.923 |
| Jrip | 0.899 | 0.900 | 0.899 |
| DT | 0.923 | 0.923 | 0.923 |
| DS | 0.868 | 0.821 | 0.830 |
| PART | 0.953 | 0.952 | 0.951 |
| OneR | 0.868 | 0.821 | 0.830 |
| J48 | 0.901 | 0.902 | 0.901 |

Table 8 shows the results of classification using different classifiers, and it demonstrates that MLP performs the most accurate classification depending on precision, recall, and F- measure. One R and DS comparatively lower results as compared to other classifiers.

## 4.9 Symmetrical Attribute Evaluation

This algorithm helps to measure the individual worth of the attribute that measures symmetrical uncertainty. It helps to detect the feature of the subset based on the training data.

*Table 9 Evaluation Performance of Symmetrical Attribute*

| Algorithm Classification | Precision | Recall | F-Measure |
|---|---|---|---|
| NB | 0.899 | 0.896 | 0.897 |
| BN | 0.901 | 0.888 | 0.891 |
| SL | 0.901 | 0.888 | 0.891 |
| MLP | 0.994 | 0.994 | 0.994 |
| NBU | 0.918 | 0.919 | 0.918 |
| SMO | 0.923 | 0.923 | 0.923 |
| Jrip | 0.899 | 0.900 | 0.899 |
| DT | 0.923 | 0.923 | 0.923 |
| DS | 0.868 | 0.821 | 0.830 |
| PART | 0.953 | 0.952 | 0.951 |
| OneR | 0.868 | 0.821 | 0.830 |
| J48 | 0.901 | 0.902 | 0.901 |
| RepT | 0.914 | 0.915 | 0.914 |

Table 9 shows the classification results of the different classifiers based on precision, recall, and F-measure. MLP shows the highest classification result, whereas DS shows the lowest classification. Results evaluated as per measuring the worth of single attribute that measures symmetrical uncertainty.

Table 10 shows the performance of the correctly classified instances based on the performance of the different algorithms along with the related classifiers. The performance of the MLP classifier shows higher results using every algorithm. The performance of the MLP classifier shows higher results using every algorithm.

*Table 10 Correctly Classified Instances*

| Feature Selection Algorithm | BN | NB | NBU | MLP | SL | SMO | DT | Jrip | OneR | PART | DS | J48 | RepT | Mean |
|---|---|---|---|---|---|---|---|---|---|---|---|---|---|---|
| CFS Subset Eval | 89.3 | 88.3 | 88.3 | 91.875 | 89.58 | 88.54 | 89.1 | 92.0 | 82.0 | 93.3 | 82.0 | 91.25 | 91.25 | 88.98 |
| Filtered Attribute Eval | 89.5 | 88.75 | 88.75 | 99.375 | 91.87 | 92.2 | 90 | 92.2 | 82.0 | 95.2 | 82.0 | 90.2 | 91.45 | 90.26 |
| Gain Ratio Attribute | 89.5 | 88.75 | 88.7 | 99.3 | 91.8 | 92.29 | 90 | 92.2 | 82.08 | 95.2 | 82.08 | 90.20 | 91.45 | 90.27 |
| Principal Component | 86.04 | 84.79 | 84.79 | 98.75 | 90.62 | 90.62 | 87.70 | 91.45 | 81.87 | 97.08 | 73.54 | 94.16 | 88.75 | 88.47 |
| Relief Attribute Eval | 89.58 | 88.75 | 88.75 | 99.91 | 91.87 | 92.29 | 90 | 92.29 | 82.08 | 95.20 | 82.08 | 90.2 | 91.45 | 90.34 |
| Correlation Based | 89.58 | 88.75 | 88.75 | 99.37 | 91.87 | 92.29 | 90 | 92.29 | 82.08 | 95.20 | 82.08 | 90.20 | 91.45 | 90.30 |
| Information Gain | 89.58 | 88.75 | 88.75 | 99.37 | 91.8 | 92.29 | 90 | 92.29 | 82.08 | 82.08 | 90.20 | 90.20 | 91.4 | 89.90 |
| One R | 89.58 | 88.75 | 88.75 | 99.37 | 91.87 | 92.29 | 90 | 92.29 | 82.08 | 95.20 | 82.0 | 90.20 | 91.45 | 90.29 |
| Symmetrical Attribute Eval | 89.58 | 88.75 | 88.75 | 99.37 | 91.87 | 92.29 | 90 | 92.29 | 82.08 | 95.20 | 82.08 | 90.20 | 91.45 | 90.30 |

## 5 Conclusion

In this research, different feature selection algorithms were used to help evaluate the performance of the educational data from Kaggle.com. Using the WEKA machine learning algorithms, each area further supported the evaluation of student performance. The results were based on the student training data set that contains different attributes. Using the different classifiers along with the based algorithms, no notable change was observed in using different classifiers, but the analysis shows that MLP classifiers perform better in combination with the different algorithms. The Relief Attribute Evaluation and Symmetrical Attribute show better results for educational data mining. In the future, student data of different sizes can be evaluated using feature selection fusion to gather the relevant results to evaluate the performance of each dataset.